\documentclass[runningheads]{llncs}
\usepackage[T1]{fontenc}
\usepackage{amsmath}
\usepackage{amssymb}
\usepackage{pdfrender}
\usepackage{graphicx,verbatim}
\usepackage{booktabs}
\usepackage{multirow}
\usepackage{graphicx}
\begin{document}
\title{HemExp: Clinically-Guided Latent Diffusion for Modeling Hematoma Expansion}
\titlerunning{HemExp: Hematoma Expansion Diffusion}
%

\author{Orhun Utku Aydin\inst{1}\thanks{Corresponding author: Orhun Utku Aydin, \email{orhunutkuaydin@gmail.com}}
Satoru Tanioka\inst{1} \and
Tzu I Chuang\inst{1} \and
Alexander Koch\inst{1} \and
Dimitrios Rallios\inst{1} \and
Marie Gultom\inst{1} \and
Begum Tahhan\inst{1} \and
Fujimaro Ishida\inst{2} \and
Dietmar Frey\inst{1} \and
Adam Hilbert\inst{1}
}  
\authorrunning{Aydin et al.}

\institute{CLAIM -- Charit\'e Lab for AI in Medicine, 
Charit\'e -- Universit\"atsmedizin Berlin, 
corporate member of Freie Universit\"at Berlin and Humboldt-Universit\"at zu Berlin, 
Charit\'eplatz 1, 10117 Berlin, Germany \\
\and
Department of Neurosurgery, Mie Chuo Medical Center, 2158-5 Myojin-cho, 514-1101, Hisai,Tsu, Japan}
  
\maketitle 
\begin{abstract}
Hematoma expansion (HE) after spontaneous intracerebral hemorrhage (ICH) is a major determinant of acute triage and treatment decisions in neurosurgical care. However, most existing methods provide either a binary expansion risk or a single follow-up volume, limiting uncertainty-aware decisions. We introduce HemExp, a clinically-guided latent diffusion model that generates patient-specific follow-up non-contrast CT images, along with segmentations of intraparenchymal and intraventricular hemorrhage. Generation is conditioned on baseline imaging, clinical variables, and an explicit expansion indicator, enabling controllable simulation of realistic clinical scenarios. HemExp uses a hemorrhage-aware multi-head variational autoencoder and models progression as the difference between baseline and follow-up latent representations with a conditional diffusion model. The model is trained on paired scans from 450 patients across multiple centers and evaluated on 107 patients from a held-out institution. HemExp produces spatial HE probability maps by generating multiple synthetic follow-up images per patient to estimate distributions of plausible follow-up hematoma volumes. Perturbing clinical inputs such as symptom-onset-to-imaging time or anticoagulant status shifts the predicted follow-up volume distribution. HemExp extends binary predictors and demonstrates robust estimation of clinically relevant outcomes in the imaging space, such as hematoma volume, intraventricular involvement, and mass effects. Overall, our results support controllable latent diffusion as a promising direction for uncertainty-aware modeling of early ICH progression. Our code is available: \url{https://github.com/orhunutkuaydin/HemExp}.


\end{abstract}
\section{Introduction}
Spontaneous intracerebral hemorrhage (ICH) is a neurological emergency with high mortality and long-term disability \cite{morotti_intracerebral_2023}. A key driver of early deterioration and poor functional outcome is hematoma expansion (HE), which occurs in roughly 20–30 percent of patients within the first hours after symptom onset. HE is assessed on paired non-contrast CT (NCCT) scans between baseline and short-term follow-up using volumetric thresholds (e.g., absolute growth $\geq$ 6 mL or relative growth $\geq$ 33 percent) on intraparenchymal hematoma (IPH) \cite{dowlatshahi_defining_2011,morotti_noncontrast_2020}. Acute therapies and triage decisions depend on the expected severity of early growth, creating substantial clinical interest in personalized forecasting of HE.  

Accurate forecasting of HE remains challenging and typically requires integrating multimodal information. Known HE predictors include clinical variables (e.g., anticoagulant use, onset-to-CT time) and NCCT/CTA markers (e.g., blend sign, black hole sign, spot sign) \cite{dowlatshahi_predicting_2016}. Existing methods formulate HE modeling as a binary classification problem and use radiomics \cite{li_NonContrast_2022}, deep learning \cite{tanioka_prediction_2024} automated segmentation and synthetic augmentation \cite{yu_Endtoend_2025}. However, these approaches do not explicitly characterize the anticipated progression of hematoma beyond estimating the likelihood of expansion. This is a critical limitation since clinical decisions frequently depend not only on whether expansion will occur, but on the location of bleeding, the plausible magnitude of growth, and secondary effects such as IVH extension, ventricular compression and mass effect. 

Motivated by these gaps, previous work has explored follow-up image synthesis to predict future NCCT appearance or hemorrhage segmentations from baseline imaging and metadata. Prior methods include predicting progression via displacement vector fields conditioned on clinical variables \cite{xiao_Intracerebral_2021}, forecasting final lesion masks \cite{yalcin_Hematoma_2024}, and generating follow-up NCCT using adversarial or transformer-augmented generative models to visualize short-term change patterns \cite{feng_Prediction_2025}. Generative synthesis has often been used primarily for data augmentation to improve HE classification rather than as a patient-specific simulator of outcomes \cite{yalcin_Hematoma_2024}. Despite progress, existing methods are commonly (i) non-probabilistic, producing a single deterministic prediction rather than a distribution over plausible futures; (ii) limited in counterfactual controllability over clinical variables; and (iii) show insufficient volumetric and clinical analysis of synthesized volumes. 

We propose HemExp, a clinical context guided latent diffusion approach for probabilistic synthesis of follow-up NCCT and aligned hematoma segmentations. HemExp (1) enables controllable follow-up synthesis via classifier-free guidance on clinical metadata and conditioning on expansion state, (2) proposes a hematoma-aware latent space that enables direct evaluation of clinically relevant imaging and volumetric endpoints, (3) estimates spatial uncertainty using voxel-wise hematoma expansion probability maps from multiple synthetic samples, (4) provides plug-and-play compatibility with existing HE predictors to translate scalar risks into clinically relevant outcomes in the imaging space.

\section{Methods}

\subsection{Problem Setup}
We model longitudinal hematoma progression from baseline to follow-up NCCT. Using baseline imaging, clinical variables, and an expansion indicator, we generate the follow-up CT slice and corresponding IPH and IVH segmentations.

For an axial slice indexed by $s$, let
$I_{\mathrm{base}}(s), I_{\mathrm{follow}}(s) \in [0,1]^{H\times W}$ denote the baseline and follow-up CT images, respectively. Let
$M_{\hat t}^{\mathrm{IPH}}(s), M_{\hat t}^{\mathrm{IVH}}(s) \in \{0,1\}^{H\times W}$
denote the IPH and IVH masks at longitudinal time $\hat t \in \{\mathrm{base}, \mathrm{follow}\}$.
We define a three-channel slice representation
\begin{equation}
U_{\hat t}(s) =
\big[
I_{\hat t}(s),\,
M_{\hat t}^{\mathrm{IPH}}(s),\,
M_{\hat t}^{\mathrm{IVH}}(s)
\big]
\in [0,1]^{3\times H\times W}.
\end{equation}

Each patient is associated with a clinical feature vector
$c \in \mathbb{R}^7$
and a binary hematoma expansion label
$y \in \{0,1\}$.
The objective is to model the conditional distribution
\begin{equation}
U_{\mathrm{follow}}(s)
\sim
p_\theta\!\left(
U_{\mathrm{follow}}(s)
\mid
U_{\mathrm{base}}(s),\, c,\, y
\right),
\end{equation}
where $p_\theta$ is the conditional generative model parameterized by $\theta$.

\subsection{Dataset and Preprocessing}

This study used a de-identified private dataset and was approved by the relevant institutional review boards. We included paired baseline and follow-up NCCT with follow-up within 24\,hours. Patients were split by institution: $450$ patients from Hospitals A--C for training and $107$ patients from held-out Hospital D for testing (Table 1).

IPH/IVH segmentations were generated semi-automatically: initial masks were obtained using a top-performing method from the MICCAI 2024 MBH-Seg Challenge, then corrected in ITK-SNAP~\cite{yushkevich_UserGuided_2019} by junior raters and refined by two medical doctors. Uncertain cases were discussed with a senior neurosurgeon.
Hematoma expansion was defined as relative IPH growth $\ge 33\%$ or absolute IPH growth $\ge 6$\,mL between baseline and follow-up.

NCCT volumes were clipped to $[0,100]$ HU and normalized to $[0,1]$. Volumes were skull-stripped with SynthStrip \cite{hoopes_SynthStrip_2022} and rigidly registered with ANTs \cite{avants_Reproducible_2011} to a common CT template. Follow-up scans were then affinely co-registered to the baseline scan in template space, and the same transforms were applied to IPH/IVH masks using nearest-neighbor interpolation. Volumes were resampled and center-cropped to $384\times384$ at $0.5\times0.5$ mm in-plane resolution. For 2D training, only axial slices with visible IPH were used. For 2.5D training, inputs comprised the target slice and adjacent axial slices at 5 mm spacing, sampled within ±1 cm of the hematoma z-extent.

\begin{table}[t]
\caption{Cohort characteristics of the train and test cohorts (median [IQR] or \%).}
\label{tab:cohort_chars}

\centering
{\fontsize{8}{9.5}\selectfont
\setlength{\tabcolsep}{4pt}
\renewcommand{\arraystretch}{1.05}
\begin{tabular}{lcc}
\hline
\textbf{Characteristic} & \textbf{Train (A--C, $n{=}450$)} & \textbf{Test (D, $n{=}107$)} \\
\hline
Age (years) & 72.0 [60.2, 80.0] & 72.0 [62.0, 82.0] \\
GCS & 14.0 [11.0, 15.0] & 15.0 [12.0, 15.0] \\
Systolic BP (mmHg) & 182.0 [160.0, 204.0] & 184.0 [161.0, 204.0] \\
Onset-to-CT (h) & 2.0 [1.0, 3.0] & 2.0 [1.0, 4.0] \\
Anticoagulant use (\%) & 10.2\% & 9.3\% \\
Antiplatelet use (\%) & 13.1\% & 23.4\% \\
Baseline IPH (mL) & 11.1 [4.7, 26.5] & 10.3 [3.9, 20.1] \\
IVH present (\%) & 39.3\% & 45.8\% \\
HE prevalence (\%) & 15.6\% & 12.1\% \\
Slice thickness (mm) & 2.0 [1.0, 3.0] & 1.0 [1.0, 1.0] \\
\hline
\end{tabular}
}
\end{table}
\subsection{Hemorrhage-Aware Multi-Head VAE}
We compress each slice representation $U_{\hat t}(s)$ using a KL-regularized variational autoencoder (MONAI AutoencoderKL~\cite{cardoso_MONAI_2022}) adapted for joint CT and hemorrhage-mask reconstruction. The encoder defines $q_\phi(z \mid U_{\hat t}(s))$ and samples a spatial latent $z_{\hat t}(s)\in\mathbb{R}^{C_z\times h_z\times w_z}$ via reparameterization, where $C_z=4$, $h_z=H/4$ and $w_z=W/4$ (i.e., $z_{\hat t}(s)\in\mathbb{R}^{4\times96\times96}$ for $384\times384$ inputs). The decoder uses a shared backbone with three $1\times1$ output heads (CT, IPH, IVH):
\begin{equation}
\hat U_{\hat t}(s)=\big[\hat I_{\hat t}(s),\hat M^{\mathrm{IPH}}_{\hat t}(s),\hat M^{\mathrm{IVH}}_{\hat t}(s)\big]=g_\psi(z_{\hat t}(s)).
\label{eq:vae_decoder}
\end{equation}

Here, $g_{\psi}$ denotes the decoder parameterized by $\psi$, which models the conditional likelihood and maps the latent code $z_{\hat{t}}(s)$ to the reconstructed CT, IPH, and IVH outputs. We use a multi-head decoder primarily to enable modality-specific supervision. We train the VAE on 2D slices by minimizing
\begin{equation}
\mathcal L_{\mathrm{VAE}}=\mathcal L_{\mathrm{CT}}+\mathcal L^{\mathrm{IPH}}_{\mathrm{seg}}+\mathcal L^{\mathrm{IVH}}_{\mathrm{seg}}+\beta\,\mathrm{KL}\!\left(q_\phi(z \mid U_{\hat t}(s))\;\|\;\mathcal N(0,I)\right),
\label{eq:vae_loss}
\end{equation}
where $\mathcal L_{\mathrm{seg}}^{\mathrm{IPH}}$ and $\mathcal L_{\mathrm{seg}}^{\mathrm{IVH}}$ use Dice + binary cross-entropy loss. The CT channel uses a pixel-wise $\mathcal L_2$ reconstruction loss, together with an LPIPS-based perceptual term $\mathcal L_{\mathrm{perc}}$~\cite{zhang_Unreasonable_2018} and a least-squares patch-adversarial term $\mathcal L_{\mathrm{LSGAN}}$~\cite{mao_Least_2017a}. We set the KL weight to $\beta=10^{-8}$. The VAE is trained progressively at resolutions of $96^2$, $192^2$, and $384^2$. All remaining loss weights and loss warm-up schedules are provided in the code repository.

\subsection{Conditional Latent Diffusion for Follow-up Progression}

\begin{figure}[t]
\centering
\includegraphics[
  width=\textwidth,
  keepaspectratio
]{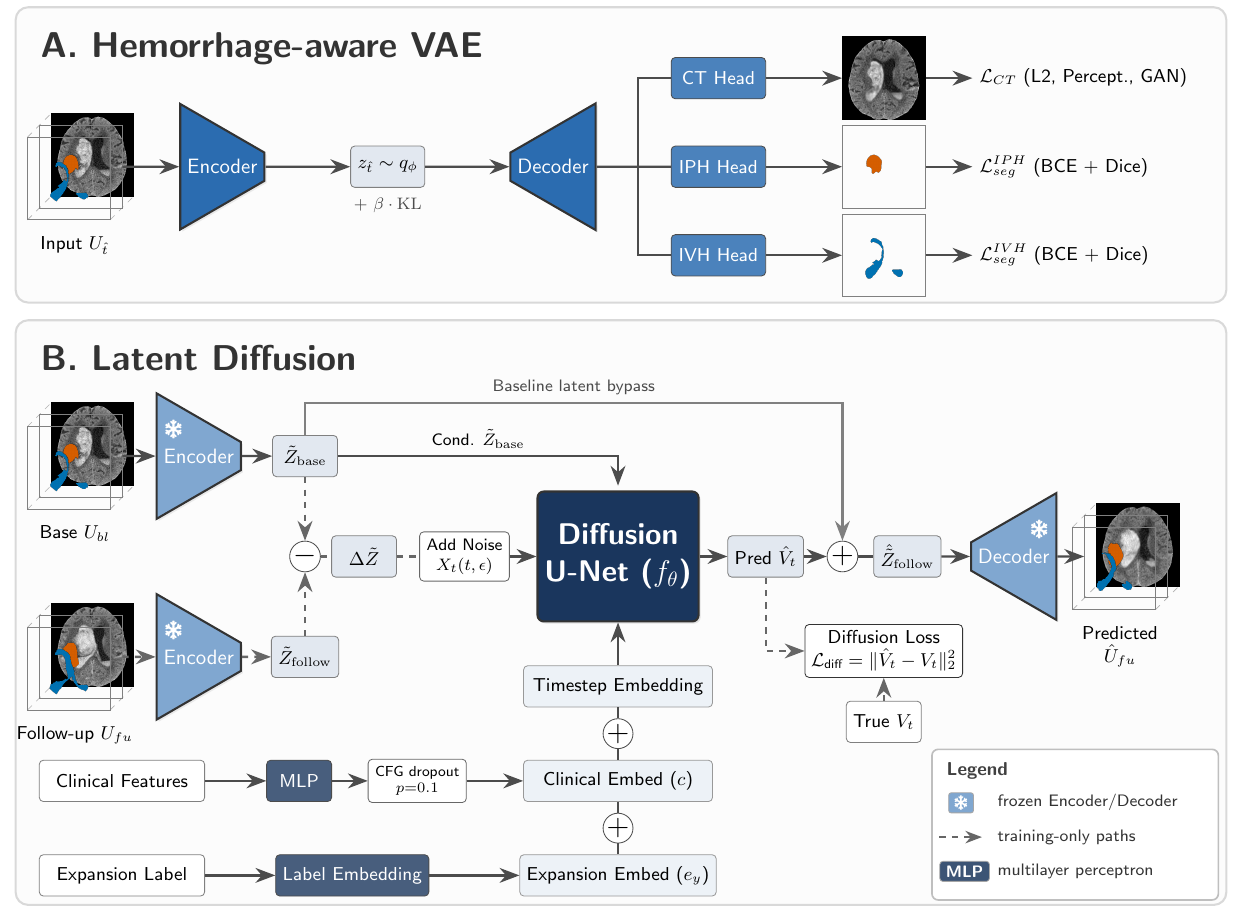}
\caption{ Overview of HemExp.}
\label{fig:figure1}
\end{figure}

Diffusion is performed in normalized latent space \cite{rombach_HighResolution_2022}, with VAE latents z-score normalized channel-wise using training-set per-channel statistics. Let $\tilde Z_{\mathrm{base}}(s)$ and $\tilde Z_{\mathrm{follow}}(s)$ denote baseline and follow-up latents. We formulate hematoma progression as a residual process in latent space,
\begin{equation}
\Delta \tilde Z(s) = \tilde Z_{\mathrm{follow}}(s) - \tilde Z_{\mathrm{base}}(s),
\end{equation}
which encourages the model to focus on progression-specific changes
while preserving subject-specific anatomy encoded in the baseline latent.

A conditional diffusion model generates $\Delta\tilde Z$ given the baseline latent and metadata. Following DDPM \cite{ho_Denoising_2020a}, let $\{\beta_t\}_{t=1}^T$ denote the variance schedule, with $\alpha_t = 1-\beta_t$ and $\bar{\alpha}_t = \prod_{i=1}^t \alpha_i$. Using $X_0:=\Delta\tilde Z$ and $t\in\{1,\dots,T\}$,
\begin{equation}
X_t=\sqrt{\bar\alpha_t}\,X_0+\sqrt{1-\bar\alpha_t}\,\epsilon,\quad \epsilon\sim\mathcal{N}(0,I).
\end{equation}
We parameterize the reverse process using the $v$-prediction formulation \cite{salimans_Progressive_2022},
\begin{equation}
V_t = \sqrt{\bar{\alpha}_t}\, \epsilon - \sqrt{1 - \bar{\alpha}_t}\, X_0,
\end{equation}
and train a conditional U-Net $f_\theta$ to predict $V_t$ as
\begin{equation}
\hat V_t = f_\theta\!\left(X_t, t \,;\, \tilde Z_{\mathrm{base}}(s), c, y \right),
\end{equation}
where conditioning is provided by the baseline latent $\tilde Z_{\mathrm{base}}(s)$,
the clinical variable vector $c$, and the expansion indicator $y$. The diffusion model is trained by minimizing
\begin{equation}
\mathcal{L}_{\mathrm{diff}}(\theta)
=
\mathbb{E}_{t\sim[1,T],\,\epsilon\sim\mathcal{N}(0,I)}
\left[
w(t)\,
\big\|
\hat V_t - V_t
\big\|_2^2
\right].
\end{equation}

where $w(t)$ denotes the Min-SNR loss reweighting \cite{hang_Efficient_2023} strategy with $\gamma = 1$.
We use a sigmoid noise schedule with $T{=}1000$ diffusion steps. At inference, we sample $\widehat{\Delta\tilde Z}$ with DDIM \cite{song_Denoising_2022}, reconstruct $\widehat{\tilde Z}_{\mathrm{follow}}=\tilde Z_{\mathrm{base}}+\widehat{\Delta\tilde Z}$, and decode with the VAE to obtain follow-up CT and IPH/IVH masks.

\subsection{Conditioning and Training}
Clinical variables $c\in\mathbb{R}^7$ (onset-to-CT time, baseline IPH volume, anticoagulant, antiplatelet, age, Glasgow Coma Scale (GCS), systolic blood pressure (SBP)) are clamped to predefined ranges, scaled to $[0,1]$, embedded by a small MLP, and added to the time embedding. The binary expansion label $y$ is encoded with a learned embedding lookup and added similarly. We apply classifier-free guidance \cite{ho_ClassifierFree_2022} only to the clinical conditioning branch, while the baseline latent and expansion-label embedding are always provided. At inference, when $y$ is unknown, we obtain it from an external multimodal HE classifier (AUC=0.82, Sensitivity=1.0, Specificity=0.62). During training, clinical conditioning is dropped with probability $0.1$ and replaced by a learned null embedding. Let $e_y$ denote the learned embedding of the binary expansion label $y$, and let $c_f := (\tilde Z_{\mathrm{base}}(s), e_y)$ denote the fixed conditioning. At sampling, guidance is applied as
\begin{equation}
\hat V_t^{\mathrm{cfg}} = \hat V_t^{\mathrm{uncond}} + \omega \left(\hat V_t^{\mathrm{cond}} - \hat V_t^{\mathrm{uncond}}\right),
\end{equation}
where $\hat V_t^{\mathrm{cond}} = f_\theta(X_t,t; c_f,c)$ and $\hat V_t^{\mathrm{uncond}} = f_\theta(X_t,t; c_f,\varnothing)$, and $\omega$ is the classifier-free guidance scale.
We optimize diffusion parameters with Adam \cite{kingma_Adam_2017} (learning rate $10^{-4}$, $\beta_1 = 0.9$, $\beta_2 = 0.99$) for $50{,}000$ steps using mixed precision, batch size 16 with gradient accumulation 2, gradient clipping 1.0, and EMA decay 0.995. To mitigate class imbalance, we sample training batches to contain equal numbers of expansion and non-expansion cases. We make our code repository available: \url{https://github.com/orhunutkuaydin/HemExp}.

\section{Results}
\begin{figure}[t]
\centering
\includegraphics[width=\textwidth,height=0.36\textheight,keepaspectratio]{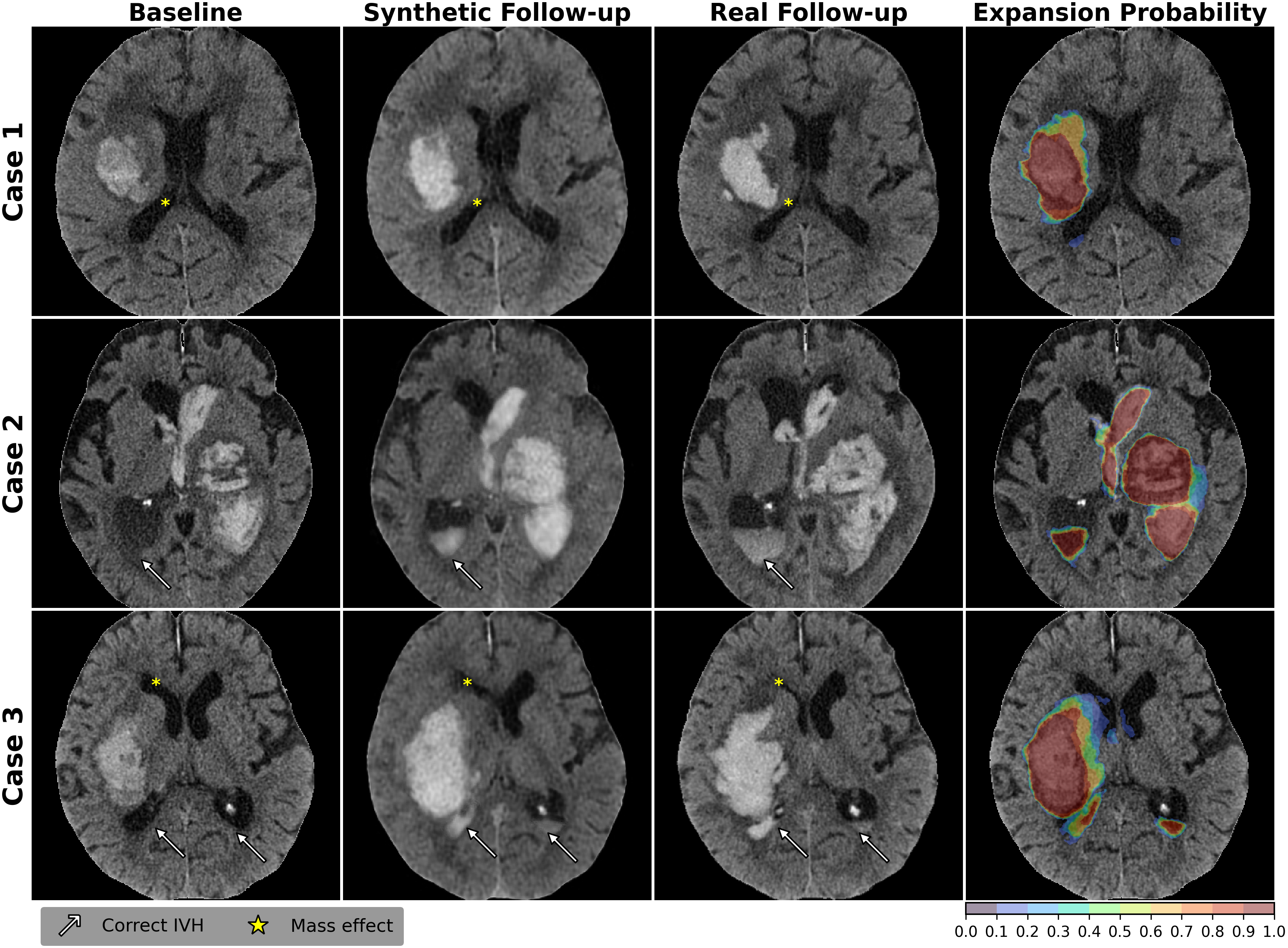}
\caption{Qualitative examples of synthetic follow-ups and expansion probability maps, showing the voxelwise fraction over 10 synthetic follow-ups labeled as hematoma.}
\label{fig:figure2}
\end{figure}
\textbf{Evaluation metrics.}
We evaluate follow-up synthesis using volumetric mean absolute error (MAE, mL), average symmetric surface distance (ASSD, mm), and hematoma coverage (COV; voxel-wise recall of the generated hemorrhage mask against the follow-up reference) for IPH and total hemorrhage. Metrics are reported stratified by expansion status.

\textbf{Inference setup.}
At inference, we sample using DDIM with 100 denoising steps and decode with the fixed VAE. Unless stated otherwise, we generate 5 samples per patient with classifier-free guidance scale $\omega=1$.

\textbf{Autoencoder fidelity.}
The VAE achieves CT SSIM, IPH Dice, and IVH Dice of 0.931/0.914, 0.986/0.986, and 0.943/0.974 (train/test), respectively.

\textbf{Follow-up NCCT generation results.} We compare HemExp variants with nnU-Net baselines while ablating conditioning from imaging only (i), imaging plus clinical variables (ic), and the expansion-status flag (ice), and evaluate through-plane context with 2.5D inputs. We also evaluate \(2.5\mathrm{D}(3)_{\mathrm{icep}}\), with expansion status provided by an external binary HE classifier (Table~2). Imaging-only HemExp is comparable to the 2D ensemble baseline, while adding the expansion status yields the most consistent gains, especially in hematoma coverage. Representative synthetic follow-ups show relevant findings, including intraventricular involvement and mass effect (Fig.~2). In a separate adherence analysis, \(2.5\mathrm{D}(3)_{\mathrm{ice}}\) matched the expansion state in \(100\%\) of cases with oracle labels, while \(2.5\mathrm{D}(3)_{\mathrm{icep}}\) simulated expansion in \(33\%\) of non-expanders with a classifier flag.

\begin{figure}[t]
\centering
\includegraphics[
  width=\textwidth,
  height=0.42\textheight,
  keepaspectratio,
  clip
]{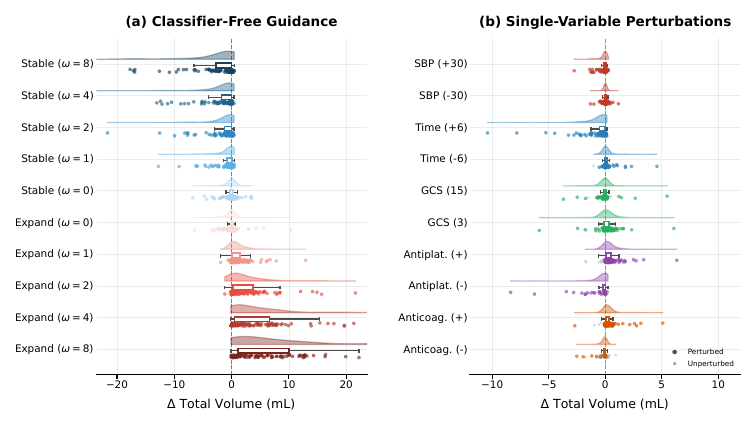}

\caption{\textbf{Clinical Controllability Analysis.}
(a) Classifier-free guidance ($\omega$) increases separation between stable and expansion profiles.
(b) Single-variable perturbations produce expected shifts in hematoma volume. SBP (mmHg) and onset time (in hours).}
\label{fig:counterfactual}
\end{figure}

\newcommand{\best}[1]{\textbf{#1}}

\begin{table*}[t]
\caption{Quantitative results. (MAE/ASSD$\downarrow$, COV$\uparrow$). i: imaging; c: clinical; e: expansion status; p: expansion flag from an external HE classifier; *: oracle expansion label at inference. 2.5D(n): Context slices. MAE in mL, ASSD in mm; values are mean(SD).}
\label{tab:quant_results_final}
\centering
\footnotesize
\renewcommand{\arraystretch}{0.95}

\begin{tabular*}{\textwidth}{@{\extracolsep{\fill}} c l @{\hspace{4pt}} ccc ccc @{} }
\toprule
& \multicolumn{1}{c}{\textbf{Method}}
& \textbf{MAE}$_{\text{IPH}}$
& \textbf{ASSD}$_{\text{IPH}}$
& \textbf{COV}$_{\text{IPH}}$
& \textbf{MAE}$_{\text{Tot}}$
& \textbf{ASSD}$_{\text{Tot}}$
& \textbf{COV}$_{\text{Tot}}$ \\
\midrule

\multirow{9}{*}{\rotatebox[origin=c]{90}{Expansion}} 
& copy-baseline            & 18.94(19.4)     & 2.72(1.7)       & 0.60(0.2)       & 28.84(22.8)     & 3.82(2.5)       & 0.56(0.2)       \\

& nnUNet-2D              & 15.96(18.2)     & 2.65(1.7)       & 0.63(0.2)       & 27.26(23.5)     & 3.16(2.4)       & 0.59(0.2)       \\

& nnUNet-3D              & 13.90(16.0)     & 2.33(1.8)       & 0.72(0.2)       & 23.95(22.2)     & 2.87(1.9)       & 0.67(0.2)       \\

& 2D$_{i}$            & 15.97(17.2)     & 2.45(1.5)       & 0.62(0.2)       & 25.89(22.4)     & 3.04(1.9)       & 0.59(0.2)       \\

& 2D$_{ic}$           & 15.48(17.0)     & 2.52(1.6)       & 0.62(0.2)       & 26.25(21.2)     & 3.16(1.8)       & 0.59(0.2)       \\

& 2D$_{ice}^{*}$         & \best{12.91}(12.8) & 2.33(1.0)       & 0.77(0.2)       & 20.24(18.4)     & 2.94(1.4)       & 0.71(0.2)       \\

& 2.5D(3)$_{ice}^{*}$     & 14.40(13.8)     & \best{2.15}(0.8) & 0.80(0.2)       & 20.17(19.5)     & 2.83(1.2)       & 0.73(0.2)       \\

& 2.5D(3)$_{ice_p}$   & 14.32(13.6)     & 2.17(0.8)       & \best{0.80}(0.2) & \best{20.03}(19.5) & 2.82(1.2)       & \best{0.73}(0.2) \\

& 2.5D(5)$_{ice}^{*}$     & 14.90(14.2)     & 2.19(0.9)       & 0.79(0.2)       & 20.61(18.5)     & \best{2.72}(1.1) & 0.73(0.2)       \\

\midrule

\multirow{4}{*}{\rotatebox[origin=c]{90}{Non-exp.}} 
& copy-baseline            & \best{1.20}(2.0) & \best{0.64}(0.3) & 0.87(0.1)       & \best{1.45}(2.1) & 1.53(2.1)       & 0.84(0.1)       \\

& nnUNet-3D            & 1.92(3.2)       & 0.66(0.4)       & \best{0.90}(0.1) & 2.03(2.8)       & 2.61(2.8)       & \best{0.87}(0.1) \\

& 2.5D(3)$_{ice_p}$   & 5.86(10.1)      & 1.12(0.8)       & 0.86(0.1)       & 6.11(10.3)      & 2.01(1.7)       & 0.83(0.1)       \\

& 2.5D(3)$_{ice}^{*}$     & 1.30(1.9)       & 0.72(0.3)       & 0.82(0.1)       & 1.46(1.8)       & \best{1.53}(1.3) & 0.79(0.1)       \\

\bottomrule
\end{tabular*}
\end{table*}

\textbf{Clinical guidance results. }We assess controllability by fixing baseline imaging and varying clinical inputs at inference. For each patient and setting we synthesize 5 follow-ups and report the mean change in total hemorrhage volume ($\Delta V_{\text{total}}$). We evaluate (i) single-variable perturbations (onset-to-imaging $\pm$6\,h, SBP $\pm$30\,mmHg, GCS set to 3/15, anticoagulant/antiplatelet toggled) and (ii) a classifier-free guidance sweep $\omega \in \{0,1,2,4,8\}$ comparing an expansion-favoring profile ($-6$\,h, +30\,mmHg, GCS=3, anticoagulant and antiplatelet use) versus a stable profile (+6\,h, SBP unchanged, GCS=15, no anticoagulant or antiplatelet use). Controllability is quantified by the separation in $\Delta V_{\text{total}}$ between profiles as $w$ increases (Fig. 3).

\section{Discussion and Conclusion}
HemExp introduces a clinical context guided latent diffusion framework for conditional simulation of short-term follow-up NCCT and hemorrhage segmentations from baseline imaging, clinical metadata, and a specified expansion state. We demonstrate clinically meaningful controllability: perturbing key clinical variables induces physiologically plausible shifts in hemorrhage volume, while classifier-free guidance improves adherence to the target clinical profile.

By generating multiple follow-up volumes per patient, HemExp enables uncertainty aware analysis through spatial hematoma expansion probability maps that localize probable growth and complement volumetric endpoints. Simulated follow-ups also capture clinically relevant image-level effects beyond scalar volume, including mass effect and intraventricular involvement. Therefore, generated follow-up images together with structured segmentation masks provide a more clinically interpretable assessment than a binary risk score alone.

Hematoma expansion forecasting remains challenging in acute, heterogeneous ICH cohorts and is further constrained by the limited size and inter-site variability of paired longitudinal CT datasets. In this setting, HemExp is best interpreted as a conditional simulator rather than a standalone predictor, and it is most reliable when the hypothesized expansion state is specified by a medical expert or provided by a validated binary predictor. 

Future work should prioritize larger multicenter longitudinal cohorts, more predictive covariates, alternative conditioning strategies, and fully 3D generation. Overall, HemExp proposes a paradigm shift in modelling hematoma expansion with potential to improve individualized clinical decision-making.

\begin{credits}
\subsubsection{\discintname}
The authors have no competing interests.  
\end{credits}

\bibliographystyle{splncs04}
\bibliography{references}
\end{document}